\documentclass[sigconf, nonacm]{acmart}
\AtBeginDocument{%
  }

\setcopyright{acmlicensed}
\copyrightyear{2018}
\acmYear{2018}
\acmDOI{XXXXXXX.XXXXXXX}
\acmConference[Conference acronym 'XX]{Make sure to enter the correct
  conference title from your rights confirmation email}{June 03--05,
  2018}{Woodstock, NY}
\acmISBN{978-1-4503-XXXX-X/2018/06}
\usepackage{subcaption}

\begin{document}

\renewcommand\floatpagefraction{.9}
\renewcommand\topfraction{.9}
\renewcommand\bottomfraction{.9}
\renewcommand\textfraction{.1}
\setcounter{totalnumber}{50}
\setcounter{topnumber}{50}
\setcounter{bottomnumber}{50}

\title{Harnessing LLM for Noise-Robust Cognitive Diagnosis in Web-Based Intelligent Education Systems}

\author{Guixian Zhang}
\email{guixian@cumt.edu.cn}
\orcid{0000-0002-7632-8411}
\affiliation{%
	\institution{School of Computer Science and Technology/School of Artificial Intelligence, China University of Mining and Technology}
	\city{Xuzhou}
	\state{Jiangsu}
	\country{China}
}

\author{Guan Yuan}
\authornote{Corresponding author.}
\email{yuanguan@cumt.edu.cn}
\orcid{0000-0003-3148-9817}
\affiliation{%
	\institution{School of Computer Science and Technology/School of Artificial Intelligence, China University of Mining and Technology}
	\city{Xuzhou}
	\state{Jiangsu}
	\country{China}
}


\author{Ziqi Xu}
\email{ziqi.xu@rmit.edu.au}
\affiliation{%
  \institution{School of Computing Technologies, RMIT University}
  \city{Melbourne}
  \state{VIC}
  \country{Australia}
}

\author{Yanmei Zhang}
\email{ymzhang@cumt.edu.cn}
\affiliation{%
	\institution{School of Computer Science and Technology/School of Artificial Intelligence, China University of Mining and Technology}
	\city{Xuzhou}
	\state{Jiangsu}
	\country{China}
}

\author{Jing Ren}
\email{jing.ren@ieee.org}
\affiliation{%
  \institution{School of Computing Technologies, RMIT University}
  \city{Melbourne}
  \state{VIC}
  \country{Australia}
}

\author{Zhenyun Deng}
\email{zd302@cam.ac.uk}
\affiliation{%
  \institution{Department of Computer Science and Technology, University of Cambridge}
  \country{United Kingdom}
}

\author{Debo Cheng}
\email{chedy055@mymail.unisa.edu.au}
\affiliation{%
 \institution{School of Computer Science and Technology, Hainan University}
	\city{Haikou}
	\state{Hainan}
	\country{China}
}

\renewcommand{\shortauthors}{Zhang et al.}

\begin{abstract}
  Cognitive diagnostics in the Web-based Intelligent Education System (WIES) aims to assess students' mastery of knowledge concepts from heterogeneous, noisy interactions. Recent work has tried to utilize Large Language Models (LLMs)  for cognitive diagnosis, yet LLMs struggle with structured data and are prone to noise-induced misjudgments. Specially, WIES's open environment continuously attracts new students and produces vast amounts of response logs, exacerbating the data imbalance and noise issues inherent in traditional educational systems. To address these challenges, we propose \textbf{DLLM}, a \textbf{D}iffusion-based \textbf{LLM} framework for noise-robust cognitive diagnosis. DLLM first constructs independent subgraphs based on response correctness, then applies relation augmentation alignment module to mitigate data imbalance. The two subgraph representations are then fused and aligned with LLM-derived, semantically augmented representations. Importantly, before each alignment step, DLLM employs a two-stage denoising diffusion module to eliminate intrinsic noise while assisting structural representation alignment. Specifically, unconditional denoising diffusion first removes erroneous information, followed by conditional denoising diffusion based on graph-guided to eliminate misleading information. Finally, the noise-robust representation that integrates semantic knowledge and structural information is fed into existing cognitive diagnosis models for prediction. Experimental results on three publicly available web-based educational platform datasets demonstrate that our DLLM achieves optimal predictive performance across varying noise levels, which demonstrates that DLLM achieves noise robustness while effectively leveraging semantic knowledge from LLM.
\end{abstract}



\received{20 February 2007}
\received[revised]{12 March 2009}
\received[accepted]{5 June 2009}

\maketitle

\section{Introduction}

The rapid advancement of the Web-based Intelligent Education System (WIES) has leveraged the pervasive interactive nature of the web~\cite{liu2023towards,liu2024inductive}, enabling personalized learning at unprecedented scale. Cognitive Diagnostic Models (CDMs) have garnered significant attention for their ability to deeply analyze students' knowledge mastery levels~\citep{liu2021towards,li2025interpretable}, making them a foundational component of the WIES. As educational data grows in volume and structural complexity, CDMs built on Graph Neural Networks (GNNs) have increasingly demonstrated their unique advantages for structured data by exploiting relational information among students, exercise, and sills~\citep{su2022graph}.

However, many existing methods are generally computed based on Identity Document (ID) without semantic knowledge~\citep{hambleton1991fundamentals,chalmers2012mirt,wang2020neural,wang2023neuralcd}, significantly constrained by the data imbalance and noise issues prevalent in web-based learning environments~\cite{guo2025enhancing}. In WIES, the response log is the record of the students' interaction with the exercises in the system and the corresponding feedback. Response logs often exhibit a long-tail distribution~\citep{wang2023self} with significant differences in behavioral patterns between high-frequency and low-frequency students. Typically, a small fraction of students complete a substantial number of online exercises, whereas most only make brief use of the platform. Existing studies~\cite{sun2022position,fu2023hyperbolic,zhang2024learning,zhang2024bayesian} indicate that GNNs-based methods tend to overemphasize high-degree nodes while neglecting low-degree nodes due to data imbalance, resulting in insufficient modeling of students with limited response logs in the WIES. 

Large language models (LLMs) leverage their internally encoded interdisciplinary knowledge systems and logical reasoning capabilities~\citep{lu2024generative,dong2025knowledge,wen2024ai} to perform in-depth analysis and contextual processing of knowledge concepts in both assessment exercises and students.
LLMs can model students' historical response logs to identify stable learning behavior patterns, common error types, and even potential learning styles. Even with limited response data, LLMs can infer deep student characteristics related to mastery levels based on external knowledge~\cite{dong2025knowledge}. On the other hand, by inferring the implicit skill requirements behind exercises from their descriptions, LLMs generate richer and more precise semantic representations of exercises. This ability to infer from external knowledge and response logs significantly enhances information richness in cognitive diagnostics.

\begin{figure}
    \centering
    \includegraphics[width=\linewidth]{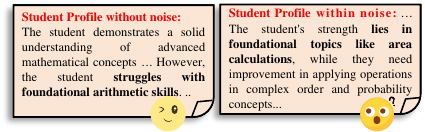}
    \caption{Student profile \#3 on the Assist0910 dataset was generated by LLM based on the original response log and the response log with 15\% added noise. The strengths have been transformed into weaknesses due to noise.}
    \label{motivation}
\end{figure}

However, relying solely on LLMs for cognitive diagnosis has significant limitations. Essentially, LLMs are text-modal general reasoning models lacking the ability to explicitly model the structured relationships inherent in web-based educational scenarios. Their reasoning heavily depends on the quality and distribution of pre-training corpora, potentially introducing external noise through hallucination phenomena~\citep{ji2023towards,du2024haloscope,10.1145/3736404}. This may cause the model to deviate from the knowledge classification systems and competency annotation standards of specific educational systems, thereby misleading diagnostic outcomes. On the other hand, when historical data contains noise such as guessing and slipping, LLM reasoning may be misled into generating incorrect conclusions. It is worth noting that WIESs generally provide simple and friendly interactions, but this also increases the possibility of noisy response logs caused by erroneous clicks. As shown in Figure~\ref{motivation}, such noise caused the LLM to generate student profiles with completely opposite semantics. Complete student profiles have been placed in Appendix A. Therefore, integrating LLMs to model structured data while eliminating noise during integration has become a critical challenge.

To address these issues, we propose a \textbf{D}iffusion-based \textbf{LLM} cognitive diagnosis framework (\textbf{DLLM}). First, to address the imbalance issue, DLLM decomposes the original response graph into correct and incorrect subgraphs, and introduce student-student edges for low-degree student nodes within each subgraph to build relation augmented subgraphs. Then, DLLM employs LLM to generate exercise descriptions and student profiles, and use their text embeddings as node representations in the semantic augmentation graph. In addition, we design a hierarchical alignment mechanism through contrastive learning. First, DLLM aligns node representations between relation-augmented subgraphs and original response subgraphs. Subsequently, the fused subgraph representations are further aligned with the semantic-augmented graph. However, noise in the data may be further amplified by data augmentation. For instance, noise in student responses may propagate through GNNs due to relation augmentation, potentially leading to erroneous inferences generated by LLMs. To achieve noise-robust representation learning, we propose a two-stage denoising diffusion module that leverages the denoising and generative advantages of the Denoising Diffusion Models (DDMs)~\cite{ho2020denoising} in the latent space. Representations in each augmented view undergo unconditional diffusion denoising followed by conditional diffusion denoising before alignment. Specifically, the unconditional DDM targets the removal of erroneous information in representations, such as guesses, slips, or misclicks in response logs and hallucinations generated by LLMs. Subsequently, a graph-guided conditional DDM is employed to eliminate misleading information, including semantic biases and information conflicts introduced by LLMs and relation augmentation. As a result, DLLM achieves noise-robust representation learning that integrates semantic knowledge with structural information. By combining with existing CMDs, it obtains the final prediction. Extensive experimental results on three real-world WIES datasets demonstrate that DLLM achieves optimal performance even under various noise conditions. The main contributions of this paper are summarized as follows:
\begin{itemize}
    \item We utilize LLM-generated student profiles and exercise descriptions to integrate semantic knowledge into CDMs, focusing on erroneous reasoning caused by noise in WIESs. To the best of our knowledge, this is the first work to consider the impact of noise on LLM cognitive diagnostic methods.
    \item We propose a \textbf{D}iffusion-based \textbf{LLM} cognitive diagnosis framework (\textbf{DLLM}), which employs a two-stage diffusion process to refine augmentation information. By utilizing node representations as conditioning, DLLM effectively eliminates noise while aligning augmentation information.
    \item We conduct extensive experiments on three datasets to evaluate the effectiveness and robustness of our framework. Experimental results demonstrate that our framework achieves state-of-the-art performance even under various noise conditions.
\end{itemize}

\section{Related Work}

In recent years, the field of cognitive diagnosis has undergone a significant evolution from traditional psychometric theoretical models to deep learning-driven approaches. Early CDMs were primarily based on Item Response Theory (IRT)~\cite{hambleton2013item}, Multidimensional IRT (MIRT)~\cite{ackerman2003using}, and the DINA~\cite{de2011generalized}. These models relied on manually designed interaction functions and strong theoretical assumptions, limiting their performance when handling large-scale response data. With the advancement of neural networks, methods such as the NCDM~\cite{wang2020neural} began utilizing neural networks to model student-problem interactions, significantly enhancing model expressiveness and prediction accuracy. CDMFKC~\cite{li2022cognitive} increases the focus on knowledge concepts by designing the difficulty and discrimination of knowledge concepts. Subsequently, a series of GNN-based CDMs have been proposed. RCD~\cite{gao2021rcd} employs graph attention mechanisms to explicitly model complex relationships among students, problems, and knowledge concepts. SCD~\cite{wang2023self} enhance representation consistency through graph self-supervised learning. KaNCD~\cite{wang2023neuralcd} improves diagnostic results for not covered knowledge concepts by modeling knowledge associations. ICDM~\cite{liu2024inductive} focuses on rapid modeling of student states within WIES, enabling inductive learning through a student-centered graph.


Although these methods excel at leveraging graph structural information, they generally overlook the heterogeneity and uncertainty present in response logs.  Furthermore, real-world noise such as guessing and slipping~\cite{pardos2011kt} further disrupts the model's inference of the student's true state. A series of graph-based cognitive diagnostic frameworks have been proposed to enhance robustness. ORCDF~\cite{qian2024orcdf} propose response-aware graph convolutional mechanisms to effectively mitigate over-smoothing in student ability representations. ISGCD~\cite{shao2025exploring} have attempted to incorporate the bottleneck principle to identify and mitigate the impact of uncertain edges in unsupervised settings, thereby enhancing diagnostic robustness to some extent.

It is noteworthy that despite LLMs demonstrating formidable knowledge representation and reasoning capabilities across multiple natural language processing tasks, their application in cognitive diagnosis remains nascent. KCD~\cite{dong2025knowledge} propose aligning LLM-generated diagnostic texts with behavioral spaces, yet face two major challenges: factual errors and semantic shifts in LLM-generated content, where such noise directly undermines the credibility of diagnostic outcomes;   and second, inherent annotation noise in raw response data further degrades LLM generation quality, rendering their prior knowledge unreliable. Consequently, effectively purifying LLM outputs and suppressing noise at both the data and model levels remains a critical issue that current research has yet to adequately address.

\section{Method}
\begin{figure*}
    \centering
    \includegraphics[width=\linewidth]{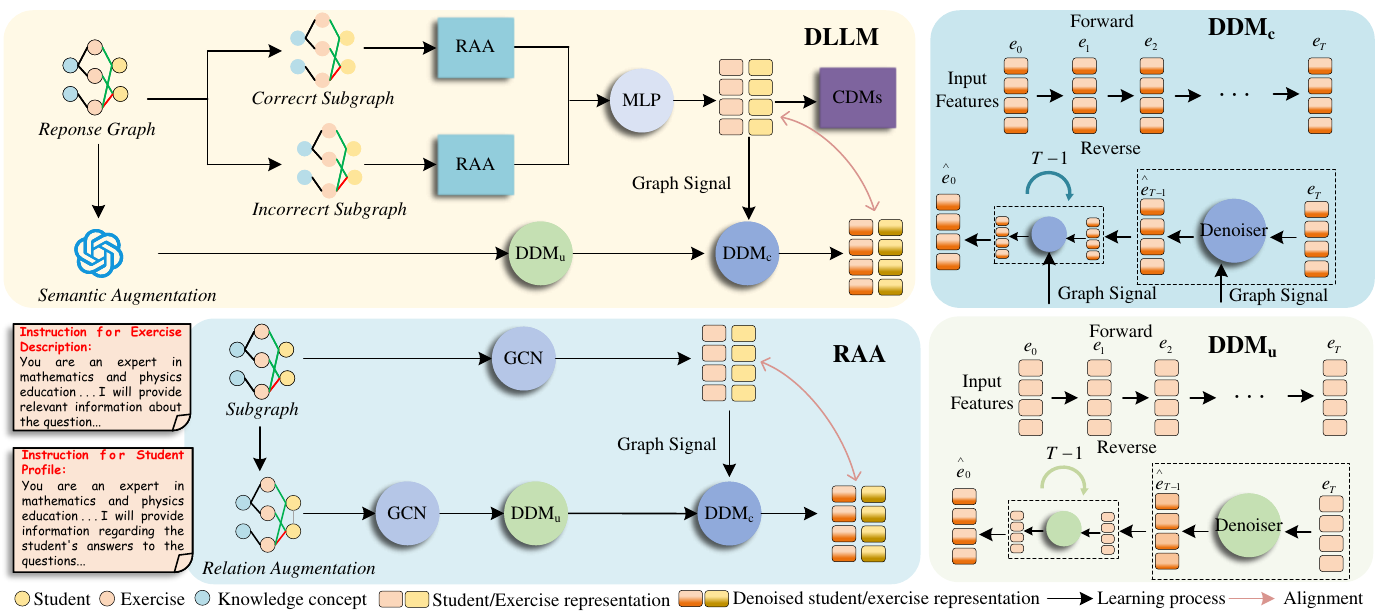}
    \caption{The proposed DLLM framework. In the figure, CDMs refers to any existing cognitive diagnostic models, RAA refers to the Relation Augmentation Alignment module, $\text{DDM}_u$ and $\text{DDM}_c$ refer to the unconditional and conditional Denoising Diffusion Model.}
    \label{framework}
\end{figure*}

The proposed Diffusion-based Large Language Model Cognitive Diagnosis (DLLM) framework comprises three core modules: Relation Augmentation Alignment, Semantic Augmentation Alignment, and a two-stage denoising diffusion module, as shown in Fig.~\ref{framework}. This framework aims to enhance collaborative information while leveraging the rich prior knowledge of LLMs to augment semantic information. It combines a graph-guided two-stage diffusion denoising model to achieve alignment of augmented information.

\subsection{Relation Augmentation Alignment}

In this section, we decompose the original response graph into correct and incorrect subgraphs based on whether the response is correct, and introduce student-student edges for low-degree student nodes within each subgraph, forming relation augmented subgraphs. In this way, we mitigate the tendency of GNNs to neglect low-degree nodes caused by  data imbalance. Then, DLLM aligns node representations between relation-augmented subgraphs and original response subgraphs through contrastive learning.


\subsubsection{Response Graph Decomposition and Relation Augmentation}
Although the response data exhibits imbalance, there is a clear distinction between completing a large number of exercises correctly and making numerous errors in the exercises. Therefore, we first decompose the interaction graph based on whether the student answered correctly. 

Given response logs $\mathcal{R} = \{(s_i, e_j, r_{ij}) \mid s_i \in S$, $e_j \in E$, $r_{ij} \in \{0,1\}\}$ where $r_{ij}=1$ denotes correct and $r_{ij}=0$ denotes incorrect, we construct a bipartite response graph $\mathcal{G} = (\mathcal{V}, \mathcal{E})$ with nodes $\mathcal{V} = S \cup E$. This graph is decomposed into two subgraphs based on response signals:  Correct Subgraph $\mathcal{G}_{\text{cor}} = (\mathcal{V}, \mathcal{E}_{\text{cor}})$ with edges $\mathcal{E}_{\text{cor}} = \{(s_i, e_j) \mid r_{ij} = 1\}$.  Incorrect Subgraph $\mathcal{G}_{\text{incor}} = (\mathcal{V}, \mathcal{E}_{\text{incor}})$ with edges $\mathcal{E}_{\text{incor}} = \{(s_i, e_j) \mid r_{ij} = 0\}$.  Adjacency matrices $\mathbf{A}_{\text{cor}} \in \{0,1\}^{\mathcal{V} \times \mathcal{V} }$ and $\mathbf{A}_{\text{incor}} \in \{0,1\}^{ \mathcal{V} \times \mathcal{V}
}$ are derived accordingly, where diagonal blocks for student-student/exercise-exercise are zero.

To enable students to obtain additional signals from neighbors, thereby alleviating the imbalance issue, we augment both subgraphs by adding edges between under-connected students. For each subgraph $\mathcal{G}_* \in \{\mathcal{G}_{\text{cor}}, \mathcal{G}_{\text{incor}}\}$, we  compute degree for each student node $s_i \in \mathcal{S}$.  We rank students by ascending degree and select bottom 50\% low-degree students: 
\begin{equation}
    S_{\text{sorted}} = \text{argsort}(\{d(s_i)\}_{i=1}^N),  
    \mathcal{S}_{\text{low}} = \{s_i \mid i \in [1, \lfloor N/2 \rfloor]\},
\end{equation} 
where $N$ denotes the number of students.

For each $s_i \in S_{\text{low}}$, randomly add an edge to a student $s_j \sim \text{Uniform}(S \setminus \{s_i\})$, generating augmented adjacency matrix $\mathbf{A}_*^{\text{aug}}$.  This yields augmented subgraphs $\mathcal{G}_{\text{cor}}^{\text{aug}}$ and $\mathcal{G}_{\text{incor}}^{\text{aug}}$, enhancing connectivity for underrepresented students.


\subsubsection{Relation Alignment}

We apply LightGCN~\citep{he2020lightgcn} to each subgraph to learn node representations. For a graph with adjacency $\mathbf{A}$, we average the aggregate information of each layer:  
\begin{equation}
    \mathbf{H} = \frac{1}{L+1} \sum_{l=0}^{L} \mathbf{H}^{(l)}_, \quad \mathbf{H}^{(l)} = \hat{\mathbf{A}} \mathbf{H}^{(l-1)}, \quad \hat{\mathbf{A}} = \mathbf{D}^{-1/2} \mathbf{A} \mathbf{D}^{-1/2}, 
\end{equation}
where $\mathbf{D}_{ii} = \sum_j \mathbf{A}_{ij}$, and $d$ is the embedding dimension. 

To align representations between the augmented subgraph and the original subgraph, we minimize an InfoNCE loss~\cite{oord2018representation} that pulls together embeddings of the same student while pushing apart different students:  
\begin{equation}
\mathcal{L}_{\text{relation}}^{stud} = -\sum_{i=1}^{N} \log \frac{\exp \left( \text{sim}(\mathbf{H}_{s_i}^{\text{ori}}, \mathbf{H}_{s_i}^{\text{aug}})/\tau \right)}{\sum_{j=1}^{N} \exp \left( \operatorname{sim}(\mathbf{H}_{s_i}^{\text{ori}}, \mathbf{H}_{s_j}^{\text{aug}})/{\tau} \right)},      
\end{equation}
where $\mathbf{H}_{s_i}$ is the embedding of student $s_i$ from a subgraph,  $\operatorname{sim}(\mathbf{u}, \mathbf{v}) = \frac{\mathbf{u}^\top \mathbf{v}}{\mathbf{u}\ \ \mathbf{v}}$ is cosine similarity,  $\tau > 0$ is a temperature hyperparameter.  This ensures invariant representation across response types.

Final student embeddings are generated by fusing correct/incorrect subgraph outputs via a Multi-Layer Perceptron (MLP):  
\begin{equation}
    \mathbf{H}_{s_i}^{\text{fused}} = \text{MLP}\left( \left[ \mathbf{H}_{s_i}^{\text{cor}} \oplus \mathbf{H}_{s_i}^{\text{incor}} \right] \right),
\end{equation} 
where $\oplus$ denotes concatenation. 

Similarly, for the exercises, we can obtain:
\begin{equation}
\mathcal{L}_{\text{relation}}^{exer} = -\sum_{j=1}^{M} \log \frac{\exp \left( \text{sim}(\mathbf{H}_{e_j}^{\text{ori}}, \mathbf{H}_{e_j}^{\text{aug}})/\tau \right)}{\sum_{m=1}^{M} \exp \left( \operatorname{sim}(\mathbf{H}_{e_j}^{\text{ori}}, \mathbf{H}_{e_m}^{\text{aug}})/ \tau \right)},
\end{equation}
\begin{equation}
    \mathbf{H}_{e_j}^{\text{fused}} = \text{MLP}\left( \left[ \mathbf{H}_{s_j}^{\text{cor}} \oplus \mathbf{H}_{s_j}^{\text{incor}} \right] \right).
\end{equation} 
Finally, we can obtain the relationship alignment loss:

\begin{equation}
    \mathcal{L}_{\text{relation}} = \mathcal{L}_{\text{relation}}^{stud} +\mathcal{L}_{\text{relation}}^{exer}.
\end{equation}
\subsection{Semantic Augmentation Alignment}


To incorporate semantic knowledge, we leverage Large Language Models (LLMs) to generate exercise descriptions and student profiles, and use the text embeddings as node representations in the semantic augmentation graph. The rich semantic information brought by LLM enables students with limited response logs to also have rich prior knowledge. Then, the fused subgraph representations are further aligned with the semantic-augmented graph through contrastive learning.

\subsubsection{Exercise Description Generation and Student Profile Generation}


In a WIES, the correct rate of questions reflects the distribution of students' mastery of specific knowledge concepts and reflects the difficulty of the question, while knowledge concepts define the cognitive dimension and semantic structure of the content. By integrating these two types of information using the LLM, we can automatically generate question descriptions that are consistent with the knowledge concepts constraints and adapt to the target difficulty level. For each exercise $e_j \in \mathcal{E}$ with associated concepts $\mathcal{K}_j$ and correct probability $p_j^{\text{cor}}$ , we generate the exercise description $T^{e}_j$ via LLM:
\begin{equation}
t^{e}_j = \text{LLM}\left( \mathcal{K}_j, p_j^{\text{cor}} \right).    
\end{equation}

By analyzing the knowledge associated with students' correct and incorrect answers, we can generate a detailed profile of their knowledge distribution and weaknesses. Correct answers reveal their mastered cognitive skills, while incorrect answers pinpoint their knowledge gaps and cognitive biases. For student $s_i \in \mathcal{S}$, we utilize LLM to generate the student profile $t^{s}_i$ based on the knowledge concepts included in each student's correct answers and incorrect answers:

\begin{equation}
    t^{s}_i = \text{LLM}\left( \mathcal{K}_i^{\text{cor}}, \mathcal{K}_i^{\text{incor}} \right).
\end{equation}

Due to space limitations, the specific instructions have been placed in Appendix B. The generated exercise description $t^{e}_j$ and student profile $t^{s}_i$ are encoded into semantic vectors:
\begin{equation}
    \mathbf{E}_j^{e} = \operatorname{Enc}(t^{e}_j), \quad \mathbf{E}_i^{s} = \operatorname{Enc}(t^{s}_i),
\end{equation}
where $\text{Enc}$ is the 
Qwen3-Embedding-0.6B model~\footnote{https://huggingface.co/Qwen/Qwen3-Embedding-0.6B}. These are projected to match the $d$-dim latent space through Principal Components Analysis (PCA):
\begin{equation}
    \mathbf{V}_{e_j} = \operatorname{PCA}(\mathbf{E}_j^{e}), \quad \mathbf{V}_{s_i} = \operatorname{PCA}(\mathbf{E}_i^{s}).
\end{equation}

\subsubsection{Semantic Alignment}

Align fused representations $\mathbf{H}^{\text{fused}}$ with LLM-generated profiles:
\begin{equation}
\mathcal{L}_{\text{semantic }}^{\text{stud}} = -\sum_{i=1}^N \log \frac{\exp \left( \text{sim}(\mathbf{H}_{s_i}^{\text{fused}}, \mathbf{v}_{s_i})/{\tau} \right)}{\sum_{n=1}^N \exp \left( \text{sim}(\mathbf{H}_{s_i}^{\text{fused}}, \mathbf{v}_{s_n})/{\tau} \right)}.
\end{equation}

Similarly for exercises:
\begin{equation}
\mathcal{L}_{\text{semantic}}^{\text{exer}} = -\sum_{j=1}^M \log \frac{\exp \left( \text{sim}(\mathbf{H}_{e_j}^{\text{fused}}, \mathbf{v}_j^{\text{exer}})/{\tau} \right)}{\sum_{m=1}^M \exp \left( \text{sim}(\mathbf{H}_{e_j}^{\text{fused}}, \mathbf{v}_m^{\text{exer}})/{\tau} \right)}.    
\end{equation}

Finally, we can obtain the relationship alignment loss:

\begin{equation}
    \mathcal{L}_{\text{semantic}} = \mathcal{L}_{\text{semantic}}^{stud} +\mathcal{L}_{\text{semantic}}^{exer}.
\end{equation}

\subsection{Two-Stage Denoising Diffusion Module}

To enhance the robustness, we propose a two-stage denoising diffusion module that combines unconditional Denoising Diffusion Model (DDM) and conditional DDM. We first employ unconditional DDM to remove erroneous information, which comprises a forward diffusion process and an unconditional denoising reverse process. Then, we conduct conditional DDM based on graph signal, which comprises a forward diffusion process and a conditional denoising reverse process.



\subsubsection{Fixed Forward Diffusion Process}

The forward diffusion process is fixed and unconditional, which is used in unconditional DDM and conditional DDM. It gradually adds Gaussian noise to the node representation $\mathbf{x}_0 = \mathbf{H}_{*} \in \mathbb{R}^{d}$ over T steps, $*$ indicates that it can be a specific student or exercise:
\begin{equation}
q(\mathbf{x}_t | \mathbf{x}_{t-1}) = \mathcal{N}(\mathbf{x}_t;   \sqrt{1-\beta_t} \mathbf{x}_{t-1}, \beta_t \mathbf{I}), \quad t = 1,\dots,T
\end{equation}
where $\beta_t \in (0,1)$ is a predefined variance schedule. We adopt a linear noise schedule where the variance parameters $\beta_t$ increase linearly from 1e-4 to 0.02 over $T$ diffusion steps to control the gradual corruption and restoration of auxiliary features. Using the reparameterization trick, we can directly sample $\mathbf{x}_t$ from $\mathbf{x}_0$:
\begin{equation}
\mathbf{x}_t = \sqrt{\bar{\alpha}_t} \mathbf{x}_0 + \sqrt{1-\bar{\alpha}_t} \epsilon, \quad \epsilon \sim \mathcal{N}(0, \mathbf{I})
\end{equation}
where $\alpha_t = 1 - \beta_t$ and $\bar{\alpha}_t = \prod_{s=1}^t \alpha_s$. This process is independent of any conditions and remains unchanged throughout.

\subsubsection{Unconditional Denoising Reverse Process}

In the first stage, we perform unconditional reverse sampling by training a denoising model to predict the added noise without external conditions. The loss function is:
\begin{equation}
\mathcal{L}_{\text{uncond}} = \mathbb{E}_{t, \mathbf{x}_0, \epsilon} \left[ \epsilon - \epsilon_\theta(\mathbf{x}_t, t) \
^2 \right],
\end{equation}
where $\epsilon_\theta$ is a MLP that predicts noise based solely on noisy input $\mathbf{x}_t$ and time step $t$.

During inference, the reverse process starts from $\mathbf{x}_T \sim \mathcal{N}(0, \mathbf{I})$ and iteratively denoises:
\begin{equation}
\mathbf{x}_{t-1} = \frac{1}{\sqrt{\alpha_t}} \left( \mathbf{x}_t - \frac{1 - \alpha_t}{\sqrt{1 - \bar{\alpha}_t}} \epsilon_\theta(\mathbf{x}_t, t) \right) + \sigma_t \mathbf{z},
\end{equation}
where $\mathbf{z} \sim \mathcal{N}(0, \mathbf{I})$ and $\sigma_t^2 = \beta_t$. 


\subsubsection{Conditional Denoising Reverse Process with Graph Signal}

In the second stage, we train an conditional DDM based on graph signal. We utilize node representations from LightGCN as conditions, allowing graph signals with structural information to guide generation. Let $\mathbf{c} = \mathbf{H}_{*} \in \mathbb{R}^{d}$ be the node representation, $*$ indicates that it can be a specific student or exercise. The conditional denoising loss is:

\begin{equation}
\mathcal{L}_{\text{cond}} = \mathbb{E}_{t, \mathbf{x}_0, \epsilon, \mathbf{c}} \left[ \epsilon - \epsilon_\theta(\mathbf{x}_t, t, \mathbf{c}) \
^2 \right].
\end{equation}

The conditioning is implemented via feature-wise linear modulation (FiLM)~\cite{perez2018film}:
\begin{equation}
\epsilon_\theta(\mathbf{x}_t, t, \mathbf{c}) = \epsilon_\theta(\mathbf{x}_t, t) \cdot (1 + \mathbf{W}_\gamma \mathbf{c}) + \mathbf{W}_\beta \mathbf{c},
\end{equation}
where $\mathbf{W}_\gamma$, $\mathbf{W}_\beta \in \mathbb{R}^{d \times d}$ are learnable projection matrices.

For conditional reverse sampling, the process is:
\begin{equation}
\mathbf{x}_{t-1} = \frac{1}{\sqrt{\alpha_t}} \left( \mathbf{x}_t - \frac{1 - \alpha_t}{\sqrt{1 - \bar{\alpha}_t}} \epsilon_\theta(\mathbf{x}_t, t, \mathbf{c}) \right) + \sigma_t \mathbf{z}
\end{equation}
This produces a refined denoised representation $\mathbf{h}^{\text{cond}}$, which leverages graph structural information for enhanced fidelity.



\subsection{Overall Training Objective}

Given input embeddings, existing Cognitive Diagnosis Models (CDMs) predict student performance on exercises as follows:

\begin{equation}
\hat{y}_{ij} = \mathcal{M}_{\text{CD}}(\mathbf{H}_{s_i}, \mathbf{H}_{e_j}, \mathbf{H}_K),
\end{equation}
where $\mathcal{M}_{\text{CD}}(\cdot)$ denotes a CDM, and $\mathbf{H}$ represents input embeddings containing representations of the student, exercise, and concepts.

To integrate DLLM with various CDMs, a transformation layer is introduced for dimension alignment. If the CDM uses a latent dimension, \textbf{H} is used directly. Otherwise, the following transformation is applied:

\begin{equation}
\mathbf{H}_t = \mathbf{H} \mathbf{W}_t + b_t, 
\end{equation}
where $\mathbf{H}_t$ serves as the input to the CDM, with trainable parameters $\mathbf{W}_t \in \mathbb{R}^{d \times Z}$ and $b_t \in \mathbb{R}^{(N+M+Z) \times 1}$. This approach allows choosing $d$ as a latent dimension, reducing the time complexity of graph convolution compared to prior methods that require $d = Z$. $Z$ represents the number of knowledge concepts.

For joint training, the binary cross-entropy (BCE) loss is used as the main objective for the cognitive diagnosis task:

\begin{equation}
\mathcal{L}_{\text{BCE}} = -\sum_{(s,e,r_{se}) \in T} \left[ r_{se} \log \hat{y}_{se} + (1 - r_{se}) \log (1 - \hat{y}_{se}) \right].
\end{equation}

The overall loss incorporates an additional consistency regularization term:

\begin{equation}
\mathcal{L} = \mathcal{L}_{\text{BCE}} + \lambda \cdot ( \mathcal{L}_{\text{relation}}+ \mathcal{L}_{\text{semantic}}+ \rho \cdot \mathcal{L}_{\text{uncond}} + \mathcal{L}_{\text{cond}}), 
\end{equation}
where $\lambda$ is a hyperparameter controlling the weight of the loss.

\section{Experiments}

In this section, we conducted comprehensive experiments to verify the predictability and robustness of DLLM.

\subsection{Setups}

\subsubsection{Datasets}

We conducted experiments based on three publicly available datasets: ASSIST0910, Junyi1808, and Eedi50. ASSIST0910 is derived from the skill-builder data~\footnote{https://sites.google.com/site/assistmentsdata/home/2009-2010-assistment-data/skill-builder-data-2009-2010} from the ASSISTments platform spanning 2009 to 2010. Junyi1808 utilizes data from the Junyi platform in August 2018~\footnote{https://www.kaggle.com/datasets/junyiacademy/learning-activity-public-dataset-by-junyi-academy}. In ASSIST0910 and Junyi1808, student response logs with fewer than 15 responses were filtered out. Eedi50 is a dataset from the NeurIPS 2020 Education Challenge, which was selected from student responses with more than 50 responses on the Eedi platform between September 2018 and May 2020.~\footnote{https://eedi.com/projects/neurips-education-challenge}.

\begin{table}[H]
\centering 
\caption{Statistics of three datasets.}
\resizebox{\linewidth}{!}{
\begin{tabular}{lccc}
\hline
Dataset    & ASSIST0910 & Junyi1808 & Eedi50  \\
\hline
\#Students   & 2,485      & 2,619        & 4,918  \\
\#Exercises   & 16,818      & 18,047        & 948    \\
\#Concepts   & 102       & 1,141         & 388    \\
\#Response Logs & 263,209    & 242,740      & 1,382,727 \\
Maximum Response Count & 1,040 & 1,076 & 827\\
Median Response Count & 44 & 52 & 239\\
Minimum Response Count & 15 & 15 & 50\\
Average Response  Count & 105.92 & 92.68 & 281.16\\
Response Log Density   & 0.0063     & 0.0051       & 0.2965  \\
Average Correctness Rate  & 0.6619     & 0.7473       & 0.5373  \\
Q Density  & 1.000      & 1.024        & 4.016     \\
\hline
\end{tabular}}
\label{data}
\end{table}

\begin{table*}[ht]
\Large
\centering 
\caption{The comparative experimental results on three datasets. \textbf{Bold} indicates the best result. "-" indicates not applicable.}
\resizebox{\textwidth}{!}{
\begin{tabular}{lcccccccccccc}
\hline
Dataset         & \multicolumn{4}{c}{ASSIST0910}                    & \multicolumn{4}{c}{Junyi1808}                     & \multicolumn{4}{c}{Eedi50}                        \\
\hline
Method          & ACC        & AUC        & F1         & DOA        & ACC        & AUC        & F1         & DOA        & ACC        & AUC        & F1         & DOA        \\
\hline
IRT             & 72.92±0.28 & 75.96±0.29 & 81.76±0.23 & -          & 77.92±0.09 & 77.28±0.20 & 86.63±0.06 & -          & 70.79±0.04 & 77.74±0.07 & 73.05±0.05 & -          \\
LightGCN-IRT    & 73.35±0.33 & 76.74±0.19 & 81.08±0.35 & -          & 78.15±0.10 & 77.59±0.12 & 86.43±0.08 & -          & 70.80±0.08 & 77.64±0.09 & 72.83±0.10 & -          \\
ORCDF-IRT       & 73.36±0.37 & 76.75±0.31 & 81.20±0.31 & -          & 78.12±0.12 & 77.22±0.43 & 86.23±0.06 & -          & 71.08±0.03 & 77.73±0.08 & 72.25±0.16 & -          \\
ISGCD-IRT       & 73.41±0.28 & 76.76±0.27 & 81.27±0.13 & -          & 78.21±0.08 & 77.88±0.15 & 86.48±0.13 & -          & 71.06±0.08 & 77.73±0.07 & 72.23±0.50 & -          \\
KCD-IRT         & 73.55±0.21 & 76.58±0.18 & 81.06±0.15 & -          & 78.73±0.05 & 77.40±0.06 & 86.93±0.07 & -          & 71.23±0.09 & 78.16±0.08 & 72.57±0.33 &            \\
DLLM-IRT        & \textbf{73.63±0.18} & \textbf{76.87±0.19} & \textbf{81.30±0.21} & -          & \textbf{79.00±0.22} & \textbf{78.12±0.35} & \textbf{87.33±0.37} & -          & \textbf{71.67±0.08} & \textbf{78.75±0.06} & \textbf{73.23±0.52} & -          \\
\hline
MIRT            & 73.25±0.33 & 76.23±0.43 & 82.11±0.32 & -          & 77.94±0.10 & 76.81±0.30 & 86.52±0.06 & -          & 71.45±0.04 & 78.36±0.07 & 72.30±0.40 & -          \\
LightGCN-MIRT   & 73.39±0.28 & 76.62±0.23 & 82.25±0.27 & -          & 78.09±0.34 & 76.96±0.49 & 86.10±0.29 & -          & 68.91±0.08 & 75.43±0.05 & 70.57±0.22 & -          \\
ORCDF-MIRT      & 73.45±0.28 & 76.92±0.25 & 81.97±0.29 & -          & 78.08±0.13 & 77.09±0.16 & 86.31±0.17 & -          & 71.74±0.06 & 78.67±0.09 & 72.53±0.60 & -          \\
ISGCD-MIRT      & 73.26±0.26 & 77.44±0.31 & 81.28±0.23 & -          & 78.25±0.15 & 77.76±0.22 & 86.54±0.06 & -          & 71.66±0.07 & 78.55±0.08 & 72.63±0.05 & -         \\
KCD-MIRT        & 73.74±0.19 & 77.77±0.17 & 80.96±0.27 & -          & 78.79±0.11 & 78.34±0.07 & 86.93±0.07 & -          & 71.79±0.07 & 78.82±0.08 & 72.78±0.21 & -          \\
DLLM-MIRT       & \textbf{73.93±0.13} & \textbf{78.01±0.13} & \textbf{81.04±0.14} & -          & \textbf{79.11±0.25} & \textbf{77.81±0.06} & \textbf{87.42±0.11} & -          & \textbf{72.00±0.13} & \textbf{78.95±0.32} & \textbf{72.86±0.27} & -          \\
\hline
NCDM            & 73.22±0.24 & 75.66±0.29 & 81.05±0.25 & 63.25±1.70 & 77.80±0.25 & 75.88±0.11 & 86.18±0.39 & 62.67±3.79 & 71.66±0.06 & 78.58±0.06 & 72.85±0.43 & 69.81±0.17 \\
LightGCN-NCDM   & 73.71±0.26 & 76.47±0.40 & 81.61±0.45 & 63.95±1.25 & 78.20±0.34 & 77.35±0.13 & 86.42±0.54 & 63.94±2.14 & 71.63±0.17 & 78.60±0.09 & 72.50±1.10 & 70.11±0.28 \\
ORCDF-NCDM      & 74.09±0.21 & 77.47±0.37 & 81.79±0.28 & 65.04±0.90 & 78.39±0.12 & 77.30±0.17 & 86.59±0.24 & 69.41±1.64 & 72.20±0.10 & 79.06±0.11 & 72.29±0.44 & 71.77±0.46 \\
ISGCD-NCDM      & 73.99±0.18 & 77.28±0.22 & 81.75±0.32 & 64.90±0.53 & 78.30±0.17 & 77.43±0.16 & 86.36±0.23 & 67.86±0.87 & 72.16±0.11 & 78.99±0.20 & 72.41±0.59 & 71.95±0.41 \\
KCD-NCDM        & 74.24±0.11 & 77.34±0.27 & 81.84±0.28 & 63.33±0.82 & 78.40±0.05 & 76.67±0.15 & 86.56±0.35 & 68.51±0.85 & 72.35±0.12 & 79.13±0.16 & 71.42±0.45 & 68.98±0.77 \\
DLLM-NCDM       & \textbf{74.46±0.18} & \textbf{77.92±0.19} & \textbf{82.06±0.21} & \textbf{65.85±0.27} & \textbf{78.91±0.13} & \textbf{77.52±0.15} & \textbf{87.31±0.25} & \textbf{69.84±0.16} & \textbf{72.77±0.12} & \textbf{79.66±0.11} & \textbf{72.91±0.44} & \textbf{73.20±0.41} \\ 
\hline
CDMFKC          & 73.00±0.25 & 75.37±0.16 & 80.86±0.40 & 63.05±1.65 & 77.58±0.14 & 75.94±0.20 &  85.87±0.24 &64.76±2.63  & 71.44±0.06 & 78.45±0.08 & 72.30±1.48 & 70.72±0.10 \\
LightGCN-CDMFKC & 73.52±0.29 & 76.71±0.23 & 81.14±0.58 & 64.25±1.14 & 77.43±0.07 & 75.22±0.20 & 85.93±0.14 & 63.96±3.61 & 71.55±0.41 & 78.21±0.61 & 72.23±0.43 & 71.86±0.40 \\
ORCDF-CDMFKC    & 73.90±0.23 & 77.47±0.28 & 81.52±0.25 & 65.56±0.33 & 78.07±0.17 & 77.32±0.11 & 86.11±0.12 & 67.54±1.15  & 72.11±0.04 & 78.92±0.06 & 72.51±0.17 & 72.70±0.15 \\
ISGCD-CDMFKC    & 73.82±0.15 & 77.25±0.24 & 81.46±0.27 & 65.07±1.18 & 78.29±0.22 & 77.39±0.26 & 86.37±0.18 & 68.38±0.96 &  72.13±0.05 & 78.94±0.07 & 72.29±0.40 & 73.25±0.21 \\
KCD-CDMFKC      & 74.08±0.14 & 77.28±0.15 & 81.60±0.23 & 64.26±0.53 & 78.52±0.15 & 77.88±0.11 & 86.56±0.17 & 67.71±0.42 & 72.13±0.19 & 78.89±0.11 & 72.68±0.33 & 69.81±0.51 \\
DLLM-CDMFKC     & \textbf{74.18±0.17} & \textbf{77.56±0.22} & \textbf{81.68±0.35} & \textbf{65.94±0.85} & \textbf{78.74±0.17} & \textbf{78.15±0.15} & \textbf{86.94±0.08} & \textbf{70.47±0.98}  & \textbf{72.58±0.13} & \textbf{79.60±0.14} & \textbf{74.13±0.23} & \textbf{73.39±0.41} \\
\hline
\end{tabular}}
\label{comp}
\end{table*}

Table~\ref{data} presents detailed statistics of these three datasets. Sparsity Rate represents the density of interactions in the student-exercise interaction matrix. Average correctness rate  reflects the average difficulty level of items across the dataset. We have compiled each student's response log counts in the table, presenting the maximum response count, median response count, minimum response count, and average response count to demonstrate the imbalance characteristics across different datasets. The datasets were partitioned into training, validation, and test sets using a 7:1:2 ratio.  For enhancing question and student representations, a generation framework based on a LLM was designed. The specific instructions will be provided in the Appendix A, and the code and datasets will be released after the acceptance of the paper.

\subsubsection{Evaluation Metrics} 
Three metrics were employed to measure classification model performance: Accuracy (ACC), Area Under the receiver operating characteristic Curve (AUC), and F1-score. Additionally, we adopt the Degree of Agreement (DOA)~\cite{qian2024orcdf} as the interpretability validation metric. The intuition behind DOA is that if student $s_a$ has higher accuracy than student $s_d$ on exercises related to concept $c_k$, then the probability of $s_a$ mastering $c_k$ should be greater, i.e., $\text{Mas}_{s_a, c_k} > \text{Mas}_{s_d, c_k}$. DOA is defined by:

\begin{equation}
 \text{DOA}_k = ( L \cdot \frac{ \sum_{j=1}^{M} Q_{jk} \, \varphi(j,a,d) \wedge \delta(r_{aj}, r_{bj}) }{ \sum_{j=1}^{M} Q_{jk} \, \varphi(j,a,d) \wedge I(r_{aj} \neq r_{bj}) } )/{ Z },    
\end{equation}
where $L = \sum_{a,d \in S} \delta\left( \text{Mas}_{s_a, c_k}, \text{Mas}_{s_d, c_k} \right)$, $Q_{jk}$ indicates the relevance of $e_j$ to $c_k$, $\varphi(j,a,d)$ checks if both students $s_a$ and $s_d$ answered $e_j$, $Z = \sum_{a,d \in S} \delta\left( \text{Mas}_{s_a, c_k}, \text{Mas}_{s_d, c_k} \right)$, $r_{aj}$ and $r_{bj}$ are the responses of $s_a$ and $s_d$ to $e_j$. Since IRT and MIRT do not use the relationship between exercises and knowledge concepts, DOA is not applicable for them.

\subsubsection{Baselines and Implementation Details}
We integrate various CDMs with different frameworks. We selected two types of CDMs as backbones: IRT~\cite{hambleton2013item} and MIRT~\cite{chalmers2012mirt} based on statistical methods, NCD~\cite{wang2020neural} and CDMFKC~\cite{li2022cognitive} based on neural networks. We adopt the hyperparameter settings and interaction
functions as described in their original papers. To fairly demonstrate our proposed DLLM framework, we combined the aforementioned CDMs with different state-of-the-art frameworks.
\begin{itemize}
    \item LightGCN~\cite{he2020lightgcn} is a powerful graph neural network algorithm that can effectively model interaction-sparse graph data while avoiding overfitting.
    \item ORCDF~\cite{qian2024orcdf} is a cognitive diagnostic framework for oversmoothing, achieving robust representation modeling through random interaction reversal and self-supervised alignment.
    \item ISGCD~\cite{shao2025exploring} is a cognitive diagnostic framework for heterogeneity and uncertainty, leveraging the information bottleneck theory to enhance graph structure and perform self-supervised alignment.
    \item KCD~\cite{dong2025knowledge} is a cognitive diagnostic framework based on LLM, utilizing a self-supervised masking mechanism to align semantic information with graph representations.
\end{itemize}


For all methods, we use Xavier to initialize parameters and Adam for optimization, the number of LightGCN layer is set to 3. For fair comparison, the embedding size $d$ of IRT and MIRT is set to 32, and $d$ of NCDM and CDMFKC is uniformly set to the number of knowledge concepts $K$. The batch size of all datasets is set to 4096. For the hyperparameters in DLLM, the learning rate is set to 0.04, $\lambda$ is searched within the range [2e-3, 1e-3, 5e-4] to find the best result, $\rho$ is set to 0.1. Each experimental result is the average under five random seeds. We adopt
Qwen-7B~\footnote{https://huggingface.co/Qwen/Qwen-7B} as the representative of LLMs. All experiments were conducted on Tesla A100 GPUs.

\begin{figure*}[htbp]
  \centering
  \begin{subfigure}{0.33\linewidth}
    \centering
    \includegraphics[width=\textwidth]{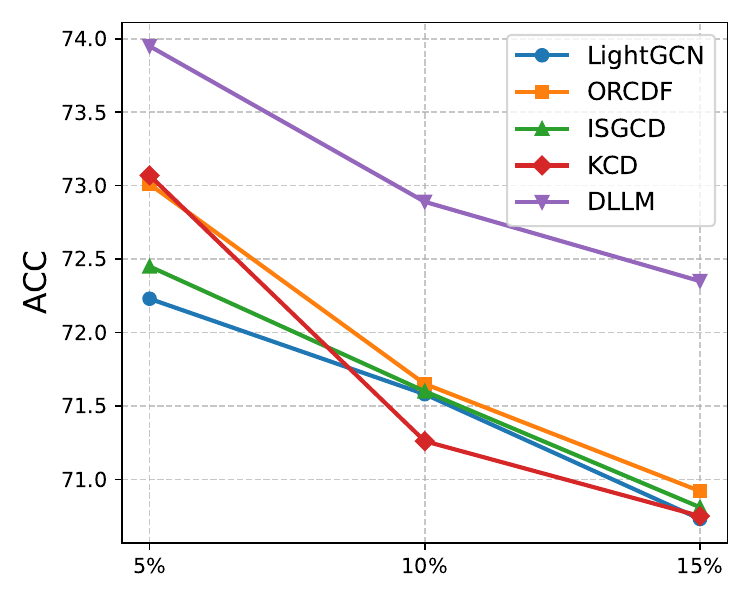} 
    \caption{ASSIST0910 dataset.}
    \label{assrobu}
  \end{subfigure}
    \begin{subfigure}{0.33\linewidth}
    \centering
    \includegraphics[width=\textwidth]{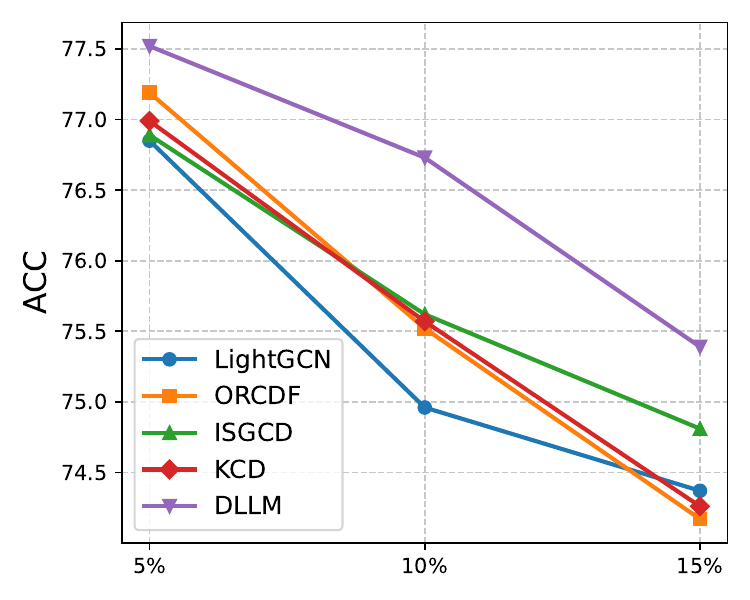} 
    \caption{Junyi1808 dataset.}
    \label{assabl}
  \end{subfigure}
  \begin{subfigure}{0.33\linewidth}
    \centering
    \includegraphics[width=\textwidth]{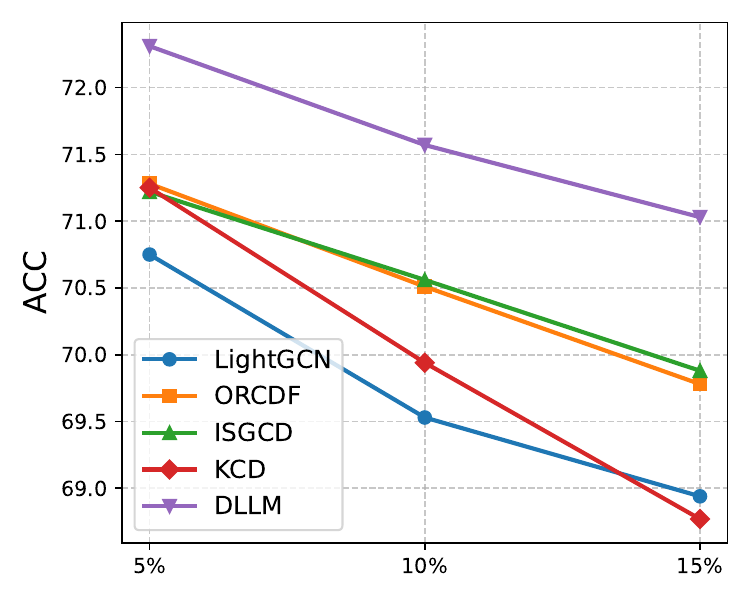} 
    \caption{Eedi50 dataset.}
    \label{junyiabl}
  \end{subfigure}
  \caption{Noise tests on three datasets under different noise conditions.}
  \label{robu}
\end{figure*}

\subsection{Comparative Experiments}

Table~\ref{comp} presents the comparative experimental results on three datasets. It can be seen that DLLM consistently outperforms baseline frameworks across all datasets and improves upon existing CDMs, demonstrating its ability to effectively suppress data noise and thereby enhance diagnostic accuracy. Specifically, MIRT models multiple latent traits of students, overcoming the limitation of traditional IRT's single-dimensional assumption, and demonstrates stronger diagnostic capability. Neural network-based models such as NCDM and CDMFKC adapt well to various representation learning frameworks and achieve strong performance. It is worth noting that under the DLLM framework, CDMFKC consistently provides better interpretability than NCDM, which benefits from the incorporation of external knowledge.

On the other hand, each framework contributes to improving CDMs performance. LightGCN uses GNNs to model a heterogeneous interaction graph and integrates existing models as a response prediction module, significantly enhancing diagnostic performance. This effectiveness stems from its ability to capture the inherent heterogeneous topological relationships among educational entities. ORCDF and ISGCD split the heterogeneous graph into a correct-response graph and an incorrect-response graph, which helps reduce confusion between correct and incorrect patterns. Both ORCDF and ISGCD employ data augmentation strategies and achieve performance comparable to LightGCN. However, ORCDF uses random response flipping during training for data augmentation. While this improves performance, it may also reinforce incorrect information present in the original data. ISGCD selectively removes edges based on uncertainty and heterogeneity measures, but this strategy risks deleting useful information, leading to situations where ISGCD performs worse than ORCDF. 

Both KCD and DLLM leverage the rich semantic knowledge of LLMs. Although KCD uses a masking strategy to partly counter potential LLM hallucinations, it does not fully address the issue of misleading inferences caused by noise. DLLM introduces a two-stage denoising mechanism that integrates relation-enhanced and semantic-enhanced alignment with denoising, leading to the best performance among all compared methods.

\subsection{Robustness Analysis}

To evaluate the robustness of various frameworks against noise, we constructed robustness tests under different noise conditions using NCDM as the backbone. Specifically, we introduced different proportions of noise based on the number of interactions in each dataset. For example, adding 5\% noise means randomly increasing 5\% of the correct response records based on the number of correct responses, while also randomly increasing 5\% of the incorrect response records based on the number of incorrect responses in the response logs. In these robustness experiments, the added noise ratios were set to 5\%, 10\%, and 15\%, respectively. The experimental results are shown in the Fig.~\ref{robu}. 

The results indicate that although the accuracy of all frameworks gradually declines as the noise level increases, they still maintain reasonable predictive performance. Crucially, the DLLM framework consistently achieves the best accuracy under all noise conditions. LightGCN, which lacks specialized anti-noise mechanisms, demonstrates relatively poor robustness. ORCDF enhances the model’s generalization ability by randomly flipping interaction types. This strategy can generate meaningful virtual negative samples in the presence of significant distribution bias, thereby helping to mitigate overfitting. However, random flipping may also reinforce incorrect information in the original data, potentially misleading model training. ISGCD prunes edges in the interaction graph based on the information bottleneck principle, but this approach carries the risk of removing useful information, which can reduce accuracy when noise is high. KCD leverages LLMs to enhance semantic information. When the noise ratio reaches 15\%, it achieves the lowest performance on the Eedi50 dataset, which has the most response records. This is because although KCD’s masking strategy can resist misinformation to some extent, the misleading information generated by the LLM due to noise is often incorrectly learned. DLLM’s two-stage denoising module effectively mitigates the impact of noise, achieving the best robustness.

\begin{figure*}[htbp]
  \centering
  \begin{subfigure}{0.33\linewidth}
    \centering
    \includegraphics[width=\textwidth]{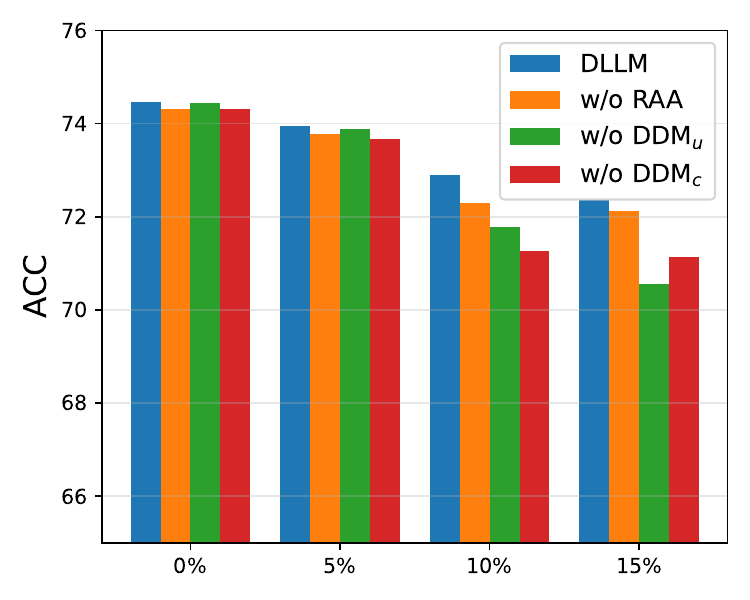} 
    \caption{ASSIST0910 dataset.}
  \end{subfigure}
    \begin{subfigure}{0.33\linewidth}
    \centering
    \includegraphics[width=\textwidth]{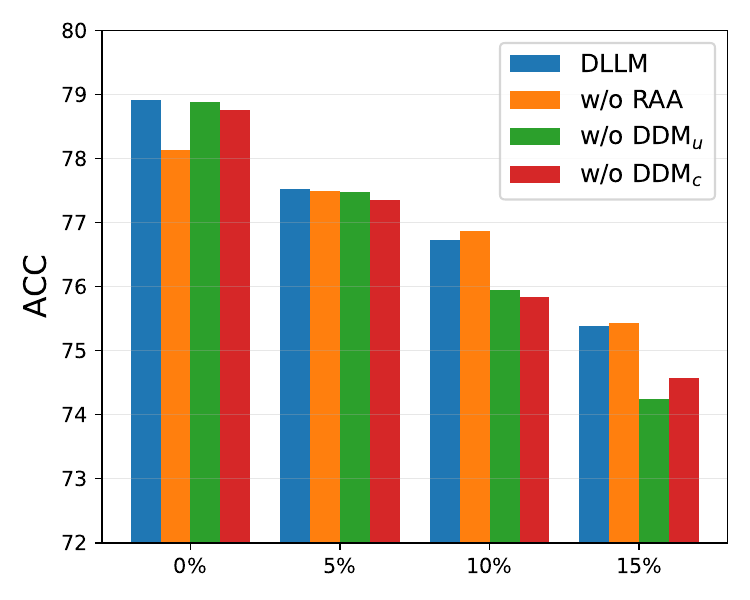} 
    \caption{Junyi1808 dataset.}
  \end{subfigure}
  \begin{subfigure}{0.33\linewidth}
    \centering
    \includegraphics[width=\textwidth]{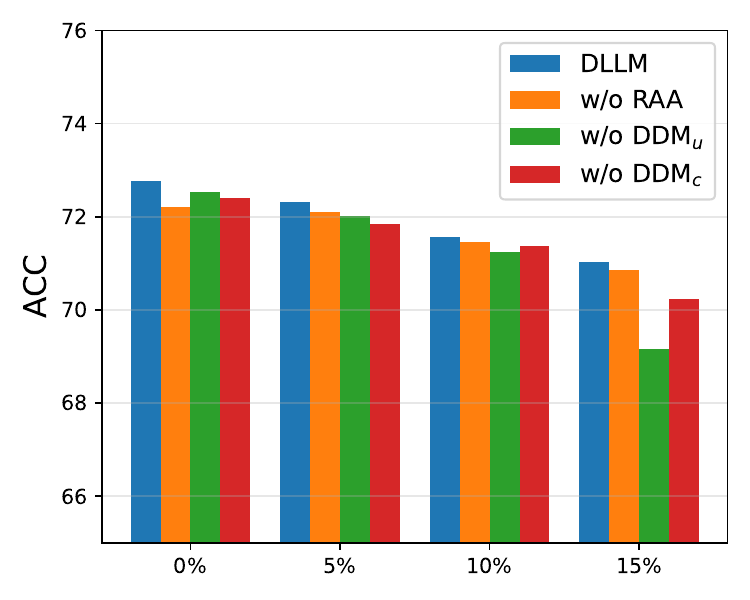} 
    \caption{Eedi50 dataset.}
  \end{subfigure}
  \caption{Ablation experiments on three datasets under different noise conditions.}
 \label{abl}
\end{figure*}

\subsection{Ablation Studies}

To clarify the contribution of each module, we conducted ablation studies on relation-enhanced alignment (w/o RAA), the unconditional denoising module (w/o DDM$_u$), and the conditional denoising module (w/o DDM$_c$), respectively. Similar to the previous robustness tests, we evaluated the performance of each variant under different noise levels. The results are shown in Fig.~\ref{abl}. The experimental results indicate that on the ASSIST0910 and Eedi50 datasets, which have relatively dense response logs, removing RAA leads to a decline in accuracy. RAA mitigates data imbalance by enhancing associations among different students, allowing low-degree nodes to receive sufficient training and thereby providing more effective graph signals for the conditional DDM. Although LLM enables students with limited interaction to acquire rich semantic knowledge, conditional DDM may erroneously eliminate some correct information if RAA is lost. However, on the sparse Junyi1808 dataset, RAA may have a negative effect under high noise levels due to the potential propagation of erroneous information.

In contrast, the unconditional DDM and conditional DDM play distinct roles under different noise conditions. On the ASSIST0910 and Junyi1808 datasets within the 0\%–10\% noise range, and on Eedi50 under 0\%–5\% noise conditions, the performance of the model with the conditional DDM removed (w/o DDM$_c$) was better than that with the unconditional DDM removed (w/o DDM$_u$). This suggests that at lower noise levels, the conditional DDM can more precisely remove noise interference based on node representations. However, as noise increases further and the node representations themselves become contaminated, the role of the unconditional DDM becomes more critical, demonstrating its stability in high-noise environments.

\subsection{Hyperparameter Analysis}

\begin{figure}
\begin{subfigure}{0.49\linewidth}
    \centering
    \includegraphics[width=\textwidth]{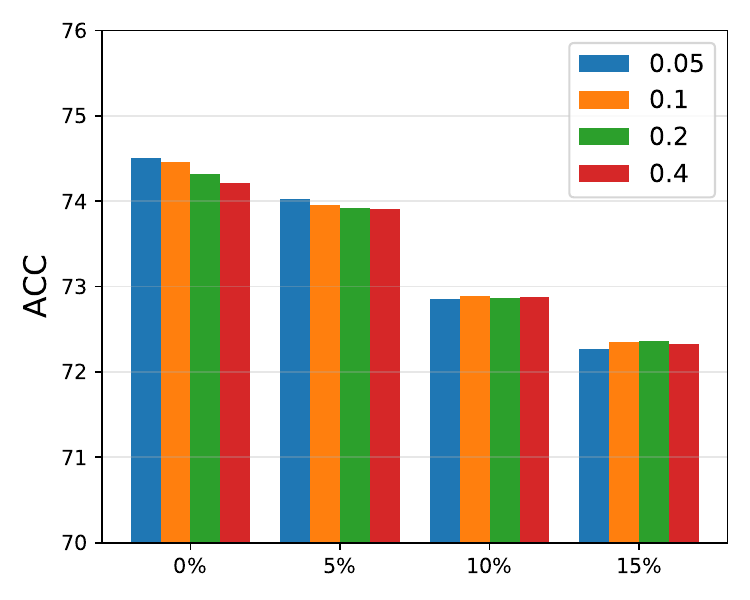} 
    \caption{ASSIST0910 dataset.}
  \end{subfigure}
\begin{subfigure}{0.49\linewidth}
    \centering
    \includegraphics[width=\textwidth]{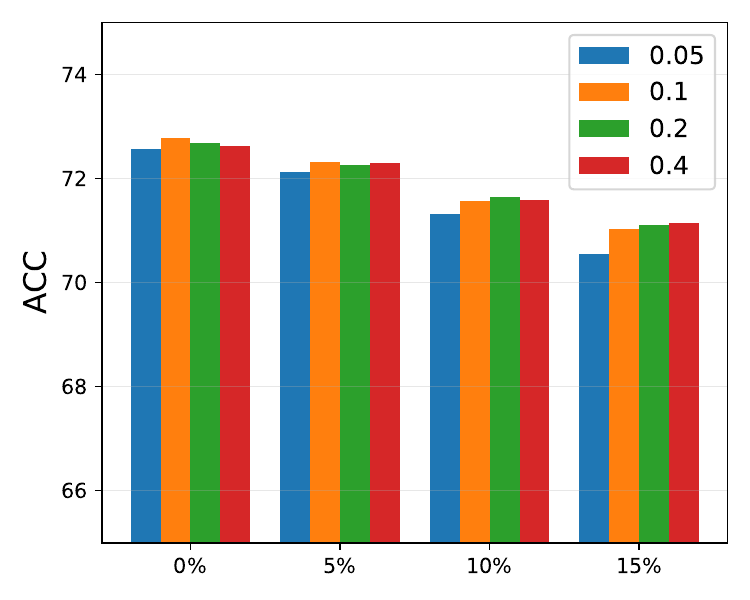} 
    \caption{Eedi dataset.}
  \end{subfigure}
    \caption{Hyperparameter analysis on two datasets.}
    \label{hyper}
\end{figure}

To investigate the collaborative relationship between the unconditional DDM and the conditional DDM, we conducted a hyperparameter analysis on the parameter  $\rho$. The experimental results on the ASSIST0910 and Eedi50 datasets are shown in the accompanying figure. Due to space constraints, the results for the Junyi1808 dataset are provided in Appendix C.

The results indicate that when $\rho$= 0.05 , the model achieved the best performance on the ASSIST0910 and Junyi1808 datasets without extra injected noise, but performed poorly on the Eedi50 dataset, which has the most interaction records. This suggests that real-world datasets inherently contain a certain level of noise, and larger datasets may include more noise. When $\rho$ was set to 0.2, the model showed stronger robustness under high-noise conditions, demonstrating the unconditional DDM’s ability to suppress noise effectively. However, when  $\rho$  was increased to 0.4, performance declined noticeably. This implies that an overly strong denoising operation, when lacking sufficient guidance from graph signals, may damage useful information in the original data and impair diagnostic accuracy.

\section{Conclusion}

Web-based intelligent education systems face increasingly prominent imbalance and noise issues due to the continuous influx of registered students and their easy-to-use interactions. In this paper, we propose a model-agnostic noise-robust cognitive diagnostic framework, DLLM. It efficiently leverages large language models (LLMs) to enhance the knowledge level of traditional cognitive diagnostic models (CDMs), and effectively addresses imbalance and noise issues through Relation Augmentation Alignment (RAA) and a two-stage denoising module. Specifically, the LLM-based semantic augmentation module endows the CDM with rich knowledge and robust reasoning capabilities. The RAA module generates node representations for the conditional denoising module that mitigate the effects of imbalance. The two-stage denoising module enhances representation robustness and facilitates alignment. The combination of unconditional DDM and conditional DDM enables effective handling of different levels of noise environments. The effectiveness and robustness of DLLM are demonstrated through datasets from three real-world web-based education platforms.



\bibliographystyle{ACM-Reference-Format}
\bibliography{sample-base}

\appendix

\section{Examples of Student Profile}

Student profile \#3 on the Assist0910 dataset was generated by LLM based on the original response log and the response log with 15\% added noise. The strengths have been transformed into weaknesses due to noise.

\subsection{Generation based on the Original Response Log}

\textbf{The student demonstrates a solid understanding of advanced mathematical concepts} such as algebraic problem-solving and complex order of operations. \textbf{However, the student struggles with foundational arithmetic skill}s, including decimal operations, fractions, and absolute value, which are critical for more complex mathematical tasks. While the student can handle higher-order problems, significant gaps in basic operations suggest the need for targeted practice and reinforcement in these fundamental areas.	The student possesses strong skills in algebraic solving and order of operations, particularly with positive reals and integers. Yet, they exhibit weaknesses in basic arithmetic operations (decimal addition and subtraction, fractions, absolute value). This imbalance indicates that while the student can handle higher-level problems, foundational gaps must be addressed to ensure comprehensive mastery.

\subsection{Generation based on the Response Log with 15\% Added Noise}

The student has demonstrated proficiency in various areas of mathematics, such as addition and subtraction of fractions, multiplication and division of integers, and solving simple algebraic equations. However, there are significant gaps in understanding probability, order of operations, and addition and subtraction of positive decimals. \textbf{The student's strength lies in foundational topics like area calculations, while they need improvement in applying operations in complex order and probability concepts.}	The student has answered most questions correctly, indicating a broad understanding of the topics. The majority of the right\_skills show a good grasp of arithmetic operations with fractions and integers, as well as some algebraic skills. However, the consistently wrong\_skills reveal gaps in probability, order of operations, and decimal arithmetic. This suggests the student needs additional support in applying mathematical operations in more complex scenarios and in probability-related problems.

\section{Details of LLM Generation}
In this section, we provide examples of the generation process of student profiles and exercise descriptions for different datasets. It is worth noting that the instructions remain the same within each example, but the input prompts change depending on the student or exercise.

\subsection{ASSIST0910}

\subsubsection{Student Profile \#0}

\textbf{Instruction:}

You are an expert in elementary mathematics education. I need your assistance in describing the characteristics of the current student, including their learning situation and mastery of knowledge concepts. I will provide information regarding the student's answers to the questions. Here are the instructions:

The information I will give you:
{"right\_num": "Number of questions answered correctly",    
"right\_skills": "Knowledge concepts included in correctly answered questions", "wrong\_num": "Number of questions answered incorrectly",     
"wrong\_skills": "Knowledge concepts included in incorrectly answered questions" }

Requirements: 

1. Please provide your decision in JSON format, following this structure: \{    "summarization": "A summarization of the student's learning situation and mastery of knowledge concepts",    "reasoning": "briefly explain your reasoning for the summarization"\}  2. Please ensure that the "summarization" is no longer than 200 words. 3. Please ensure that the "reasoning" is no longer than 200 words. 4. Do not provide any other text outside the JSON string. 

Here is the information:

\textbf{Input Prompt:}

\{"right\_num": 33, "right\_skills": ["Area Trapezoid", "Area Trapezoid", "Area Trapezoid", "Area Trapezoid", "Area Trapezoid", "Area Trapezoid", "Area Trapezoid", "Area Trapezoid", "Area Trapezoid", "Area Trapezoid", "Area Trapezoid", "Area Trapezoid", "Area Trapezoid", "Area Trapezoid", "Area Trapezoid", "Area Trapezoid", "Area Trapezoid", "Area Trapezoid", "Area Trapezoid", "Area Trapezoid", "Area Trapezoid", "Area Trapezoid", "Area Trapezoid", "Circumference ", "Circumference ", "Area Circle", "Area Circle", "Area Trapezoid", "Area Trapezoid", "Area Trapezoid", "Circumference ", "Circumference ", "Circumference "], 

"wrong\_num": 18, "wrong\_skills": ["Area Trapezoid", "Area Trapezoid", "Area Trapezoid", "Area Trapezoid", "Area Trapezoid", "Area Trapezoid", "Area Trapezoid", "Circumference ", "Circumference ", "Circumference ", "Circumference ", "Circumference ", "Area Circle", "Area Circle", "Area Circle", "Area Trapezoid", "Circumference ", "Circumference "]\}

\subsubsection{Exercise Description \#0}

\textbf{Instruction:}

You are an expert in elementary mathematics education. I need your assistance in describing the type of question at hand and outlining the skills students should possess to solve it correctly. I will provide relevant information about the question, which is from the subject of elementary mathematics. Here are the instructions:

1. I will provide the question description information in JSON string format: {    "skill": "The skill name relevant to this question",    "question\_diff": "The statistical difficulty of the question, represented by the overall correct rate of all students who answered it"}   

Requirements: 1. Please provide your answer in JSON format, following this structure: \{"summarization": "summary description of the question type",    "reasoning": "briefly explain your reasoning for the summarization"\} 2. Please ensure that the "summarization" is no longer than 200 words. 3. Please ensure that the "reasoning" is no longer than 200 words. 4. Do not provide any other text outside the JSON string. Here is the information: 

\textbf{Input Prompt:}

{	"skill":"Estimation",	

"question\_diff":"2/4"}

\subsection{Junyi1808}

\subsubsection{Student Profile \#0}

\textbf{Instruction:}

You are an expert in elementary mathematics education. I need your assistance in describing the characteristics of the current student, including their learning situation and mastery of knowledge concepts. I will provide information regarding the student's answers to the questions. Here are the instructions: 

The information I will give you:\{    "right\_num": "Number of questions answered correctly",   "right\_skills": "Knowledge concepts included in correctly answered questions",    "wrong\_num": "Number of questions answered incorrectly", "wrong\_skills": "Knowledge concepts included in incorrectly answered questions"\}

Requirements:1. Please provide your decision in JSON format, following this structure: { "summarization": "A summarization of the student's learning situation and mastery of knowledge concepts",    "reasoning": "briefly explain your reasoning for the summarization"} 2. Please ensure that the "summarization" is no longer than 200 words. 3. Please ensure that the "reasoning" is no longer than 200 words. 4. Do not provide any other text outside the JSON string. 

Here is the information:

\textbf{Input Prompt:}
\{"right\_num": 38, "right\_skills": ["[ Basic ] One on one correspondence counting", "[ Basic ] Size comparison within 10", "[ Fundamentals ] 6 and 6", "[ Basic ] One on one correspondence counting", "[Basic] Which one is it", "[ Basic ] One on one correspondence counting", "[Basic] Which one is it", "[ Basic ] Numbers from 11 to 20", "[ Basic ] Numbers from 11 to 20", "[ Basic ] One on one correspondence counting", "[ Basic ] Numbers from 11 to 20", "[ Basic ] Size comparison within 10", "[ Fundamentals ] 6 and 6", "[ Fundamentals ] 6 and 6", "[ Basic ] One on one correspondence counting", "[ Basic ] One on one correspondence counting", "[ Basic ] One on one correspondence counting", "[Basic] Which one is it", "[ Basic ] Size comparison within 10", "[ Fundamentals ] 6 and 6", "[ Basic ] One on one correspondence counting", "[ Basic ] Size comparison within 10", "[ Basic ] Numbers from 11 to 20", "[ Basic ] Numbers from 11 to 20", "[ Basic ] One on one correspondence counting", "[ Fundamentals ] 6 and 6", "[ Basic ] One on one correspondence counting", "[ Basic ] One on one correspondence counting", "[ Basic ] Numbers from 11 to 20", "[ Basic ] Size comparison within 10", "[ Basic ] Size comparison within 10", "[ Basic ] Numbers from 11 to 20", "[ Basic ] Numbers from 11 to 20", "[Basic] Which one is it", "[ Basic ] One on one correspondence counting", "[ Basic ] Numbers from 11 to 20", "[ Basic ] One on one correspondence counting", "[ Basic ] Numbers from 11 to 20"], 

"wrong\_num": 15, "wrong\_skills": ["[ Fundamentals ] 6 and 6", "[ Basic ] Numbers from 11 to 20", "[ Basic ] Numbers from 11 to 20", "[ Basic ] Numbers from 11 to 20", "[Basic] Which one is it", "[ Basic ] One on one correspondence counting", "[ Basic ] Numbers from 11 to 20", "[ Basic ] Numbers from 11 to 20", "[ Basic ] Numbers from 11 to 20", "[ Basic ] Numbers from 11 to 20", "[ Basic ] Numbers from 11 to 20", "[ Fundamentals ] 6 and 6", "[ Basic ] Numbers from 11 to 20", "[ Basic ] Size comparison within 10", "[ Basic ] Numbers from 11 to 20"]\}

\subsubsection{Exercise Description \#0}

\textbf{Instruction:}
You are an expert in elementary mathematics education. I need your assistance in describing the type of question at hand and outlining the skills students should possess to solve it correctly. I will provide relevant information about the question, which is from the subject of elementary mathematics. Here are the instructions: 

1. I will provide the question description as a JSON string: \{    "skills": "Skill names relevant to this question", (separate multiple skills with semicolons)    "skills\_diff": "Difficulty level of associated skills" (separate multiple levels with semicolons, corresponding one-to-one with the skills above)   "question\_diff": "The statistical difficulty of the question, represented by the overall correct rate of all students who answered it"\}

Requirements: 1. Please provide your answer in JSON format, following this structure:\{    "summarization": "summary description of the question type",    "reasoning": "briefly explain your reasoning for the summarization"\} 2. Please ensure that the "summarization" is no longer than 200 words. 3. Please ensure that the "reasoning" is no longer than 200 words. 4. Do not provide any other text outside the JSON string. Here is the information:

\textbf{Input Prompt:}

\{"skills":"[ Basic ] One on one correspondence counting",

"skills\_diff":"easy",	

"question\_diff":"127/141"\}

\subsection{Eedi50}

Since there are more response logs for student \#0 in the Eedi50 dataset, and considering space limitations, we choose student \#3 as an example.

\subsubsection{Student Profile \#3}

\textbf{Instruction:}
You are an expert in mathematics and physics education. I need your assistance in describing the characteristics of the current student, including their learning situation and mastery of knowledge concepts.
I will provide information regarding the student's answers to the questions.
Here are the instructions:
The information I will give you:{   "right\_num": "Number of questions answered correctly",    "right\_skills": "Knowledge concepts included in correctly answered questions",    "wrong\_num": "Number of questions answered incorrectly",    "wrong\_skills": "Knowledge concepts included in incorrectly answered questions"}
Requirements:
1. Please provide your decision in JSON format, following this structure:
{    "summarization": "A summarization of the student's learning situation and mastery of knowledge concepts",    "reasoning": "briefly explain your reasoning for the summarization"\}
2. Please ensure that the "summarization" is no longer than 200 words.
3. Please ensure that the "reasoning" is no longer than 200 words. 4. Do not provide any other text outside the JSON string.

Here is the information:

\textbf{Input Prompt:}

\{"right\_num": 24, "right\_skills": ["Maths; Number; Factors, Multiples and Primes; Multiples and Lowest Common Multiple", "Maths; Number; Negative Numbers; Multiplying and Dividing Negative Numbers", "Maths; Number; Basic Arithmetic; Mental Multiplication and Division", "Maths; Number; Basic Arithmetic; Place Value", "Maths; Number; Basic Arithmetic; Mental Multiplication and Division", "Maths; Number; Basic Arithmetic; Mental Multiplication and Division", "Maths; Number; Basic Arithmetic; Mental Multiplication and Division", "Maths; Number; Basic Arithmetic; Mental Multiplication and Division", "Maths; Number; Factors, Multiples and Primes; Factors and Highest Common Factor", "Maths; Number; Basic Arithmetic; Mental Multiplication and Division", "Maths; Number; Basic Arithmetic; Place Value", "Maths; Geometry and Measure; Units of Measurement; Time", "Maths; Number; Factors, Multiples and Primes; Multiples and Lowest Common Multiple", "Maths; Number; Basic Arithmetic; Mental Multiplication and Division", "Maths; Number; Basic Arithmetic; Place Value", "Maths; Number; Basic Arithmetic; Mental Multiplication and Division", "Maths; Number; Basic Arithmetic; Mental Multiplication and Division", "Maths; Geometry and Measure; Units of Measurement; Time", "Maths; Geometry and Measure; Units of Measurement; Time", "Maths; Number; Indices, Powers and Roots; Squares, Cubes, etc", "Maths; Number; Negative Numbers; Multiplying and Dividing Negative Numbers", "Maths; Geometry and Measure; Units of Measurement; Time", "Maths; Geometry and Measure; Units of Measurement; Time", "Maths; Number; Factors, Multiples and Primes; Factors and Highest Common Factor"],

"wrong\_num": 35, "wrong\_skills": ["Maths; Number; Basic Arithmetic; Mental Multiplication and Division", "Maths; Number; Factors, Multiples and Primes; Factors and Highest Common Factor", "Maths; Geometry and Measure; Units of Measurement; Time", "Maths; Number; Indices, Powers and Roots; Squares, Cubes, etc", "Maths; Geometry and Measure; Units of Measurement; Time", "Maths; Geometry and Measure; Units of Measurement; Time", "Maths; Number; Basic Arithmetic; Mental Multiplication and Division", "Maths; Number; Basic Arithmetic; Place Value", "Maths; Number; Basic Arithmetic; Place Value", "Maths; Number; Basic Arithmetic; Mental Multiplication and Division", "Maths; Number; Basic Arithmetic; Mental Multiplication and Division", "Maths; Number; Factors, Multiples and Primes; Multiples and Lowest Common Multiple", "Maths; Number; Basic Arithmetic; Mental Multiplication and Division", "Maths; Geometry and Measure; Units of Measurement; Time", "Maths; Number; Basic Arithmetic; Mental Multiplication and Division", "Maths; Number; Factors, Multiples and Primes; Multiples and Lowest Common Multiple", "Maths; Geometry and Measure; Units of Measurement; Time", "Maths; Number; Indices, Powers and Roots; Squares, Cubes, etc", "Maths; Number; Factors, Multiples and Primes; Factors and Highest Common Factor", "Maths; Number; Indices, Powers and Roots; Squares, Cubes, etc", "Maths; Number; Basic Arithmetic; Place Value", "Maths; Geometry and Measure; Units of Measurement; Time", "Maths; Number; Basic Arithmetic; Place Value", "Maths; Number; Basic Arithmetic; Mental Multiplication and Division", "Maths; Geometry and Measure; Units of Measurement; Time", "Maths; Number; Basic Arithmetic; Place Value", "Maths; Number; Basic Arithmetic; Mental Multiplication and Division", "Maths; Number; Factors, Multiples and Primes; Prime Numbers and Prime Factors", "Maths; Number; Basic Arithmetic; Mental Multiplication and Division", "Maths; Number; Basic Arithmetic; Mental Multiplication and Division", "Maths; Number; Basic Arithmetic; Mental Multiplication and Division", "Maths; Number; Factors, Multiples and Primes; Factors and Highest Common Factor", "Maths; Number; Basic Arithmetic; Place Value", "Maths; Number; Basic Arithmetic; Mental Multiplication and Division", "Maths; Number; Factors, Multiples and Primes; Factors and Highest Common Factor"]\}

\subsubsection{Exercise Description \#0}

\textbf{Instruction:}

You are an expert in mathematics and physics education. I need your assistance in describing the type of question at hand and outlining the skills students should possess to solve it correctly.
I will provide relevant information about the question, which is from the subjects of mathematics and physics.
Here are the instructions:
1. I will provide the question description information in JSON string format:\{    "skills": "Skill names relevant to this question", (separate multiple skills with semicolons)    "question\_diff": "The statistical difficulty of the question, represented by the overall correct rate of all students who answered it"\}

Requirements:
1. Please provide your answer in JSON format, following this structure:\{    "summarization": "summary description of the question type",    "reasoning": "briefly explain your reasoning for the summarization"\}
2. Please ensure that the "summarization" is no longer than 200 words.
3. Please ensure that the "reasoning" is no longer than 200 words.
4. Do not provide any other text outside the JSON string.
Here is the information:

\textbf{Input Prompt:}

{"skills":"Maths; Algebra; Writing and Simplifying Expressions; Writing Expressions",	

"question\_diff":"1138/2162"}

\section{Hyperparameter Analysis}

Considering space limitations, we place the Junyi1808 dataset here. For the corresponding hyperparameter analysis, please refer to Section 4.4.

\begin{figure}[h]
    \centering
    \includegraphics[width=\linewidth]{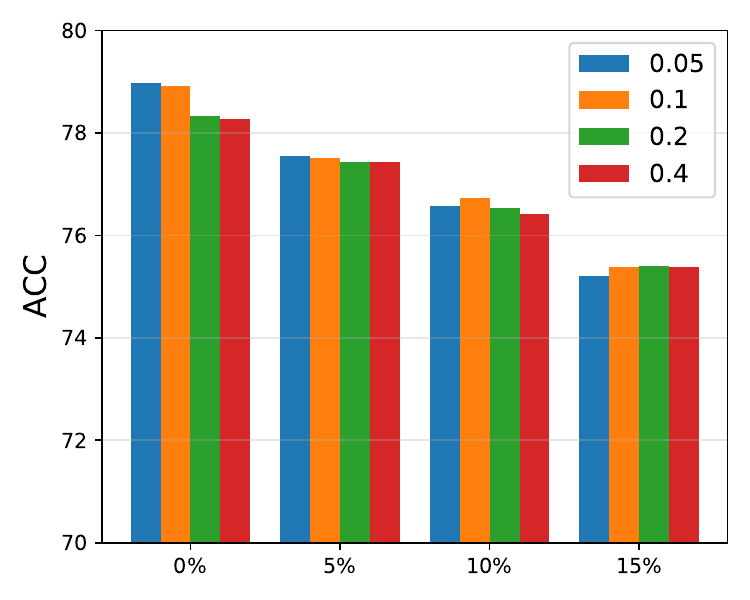}
    \caption{Hyperparameter Analysis on the Junyi1808 dataset.}
    \label{hyj}
\end{figure}

\end{document}